\documentclass[submission,copyright,creativecommons]{eptcs}
\providecommand{\event}{QFM2012} 
  \usepackage{breakurl}

\usepackage{amsmath}
\usepackage{multirow,rotating}
\usepackage{amsfonts}
\usepackage{xspace}
\usepackage{algorithm}               
\usepackage{algorithmic}             
\usepackage{psfrag}
\usepackage{array}
\usepackage{url}
\usepackage{amssymb}
\usepackage{hyperref}
\usepackage[usenames]{color}
\usepackage[numbers, sort&compress]{natbib}
\usepackage{subfigure}

\usepackage[none]{hyphenat}

\newtheorem{defn}{Definition}

\newcommand{\Nset}{\mathbb N}

\newcommand{\until}{\mbox{\sf U}}

\newcommand{\candS}{\mbox{$ S$}}
\newcommand{\sizeof}[1]{\mid\! #1 \!\mid}

\newcommand{\strings}{\emph{strings}}

\newcommand{\mdp}{\mbox{$  M$}}

\newcommand{\sch}{\mbox{$\mathfrak{S}$}}

\newcommand{\RED}{\textsc{RED}\xspace}
\newcommand{\BLUE}{\textsc{BLUE}\xspace}
\newcommand{\IOFPTA}{{\sc IOFPTA}}

\newcommand{\prefix}{\mathrm{prefix}}
\newcommand{\pmin}{\mathrm{min}}
\newcommand{\pmax}{\mathrm{max}}
\newcommand{\COMPATIBLE}{\emph{Compatible}}

\newcommand{\alergia}{{\sc Alergia}}
\newcommand{\ioalergia}{{\sc IOalergia}}

\newcommand{\MERGE}{\emph{Merge}}
\newtheorem{example}{Example}
\newcommand{\seq}{\mathrm{Seq}}
\newcommand{\tree}{\mathrm{Tree}}
\newcommand{\tran}{\mathrm{Tran}}
\newcommand{\tabincell}[2]{\begin{tabular}{@{}#1@{}}#2\end{tabular}}

\begin{document}

%
\title{Learning Markov Decision Processes\\ for Model Checking}
\author{Hua Mao, Yingke Chen, Manfred Jaeger, Thomas D. Nielsen, Kim G. Larsen, and Brian Nielsen
\institute{Department of Computer Science\\
Aalborg University\\ Denmark}
\email{[huamao,ykchen,jaeger,tdn,kgl,bnielsen] @cs.aau.dk}}
\def\authorrunning{H. Mao, Y. Chen, M. Jaeger, T. D. Nielsen, K. G. Larsen, \& B. Nielsen}
\def\copyrightholders{H. Mao, Y. Chen, M. Jaeger, T. D. Nielsen, \\ K. G. Larsen, \& B. Nielsen}
\def\titlerunning{Learning Markov Decision Processes for Model Checking}
\maketitle

\begin{abstract}
Constructing an accurate system model for formal model verification can be both resource demanding and time-consuming. To alleviate this shortcoming, algorithms have been proposed for automatically learning system models based on observed system behaviors. In this paper we extend the algorithm on learning probabilistic automata to reactive systems, where the observed system behavior is in the form of alternating sequences of inputs and outputs. We propose an algorithm for automatically learning a deterministic labeled Markov decision process model from the observed behavior of a reactive system. The proposed learning algorithm is adapted from algorithms for learning deterministic probabilistic finite automata, and extended to include both probabilistic and nondeterministic transitions. The algorithm is empirically analyzed and evaluated by learning system models of slot machines. The evaluation is performed by analyzing the probabilistic linear temporal logic properties of the system as well as by analyzing the schedulers, in particular the optimal schedulers, induced by the learned models.
\end{abstract}

\section{Introduction}
\label{sec:introduction}
Model checking is successfully used in many areas to check a formal system model against a specification given by a logical expression. However, to construct an accurate model of an industrial system is usually difficult and time consuming. The difficulty of model construction, or system modeling, is regarded by industry as a challenge to adopt other powerful model-driven development (MDD) techniques and tools as well. Meanwhile, the necessary accurate, updated and detailed documentations rarely exist for legacy software or 3rd party components. Therefore, we consider system model learning techniques~\cite{RafSteBer05,Leucker06, SVA04a,MCJN+11}, which can automatically construct or \emph{learn} an accurate high-level model from observations of a given black-box embedded system component. Afterwards, given a learned and explicitly represented model, model checking and other MDD techniques can be applied with other existing component models.

For learning non-probabilistic system models, Angluin's approaches~\cite{Angluin87} has been well developed and implemented~\cite{Leucker06,RafSteBer05,AartsV10}. However, a disadvantage of those system models is that complex systems are often only partially observable via their interactions with the user. Even worse, the observation is often not noise-free. Compared with deterministic models, probabilistic models are more feasible to model a complicated real system and its physical components, unpredictable user interactions and the usage of randomized algorithms. In this paper, we focus on probabilistic models. Sen et al.~\cite{SVA04a} adapted the algorithm from~\cite{CarOnc94} for learning Markov chain models in purpose of verification. In \cite{MCJN+11}, a learning approach related to~\cite{SVA04a} is developed, and strong theoretical and experimental consistency results are established. Considering a limited situation that the target system is not fully under control and only a \emph{single} observation sequence is available, the algorithm for learning variable order Markov chains~\cite{RS1996} is developed to verify stationary system properties on the learned models~\cite{CMJN+12}.




In Markov chains, probabilistic choices may serve to model and quantify possible outcomes of randomized actions or the interface between a system and its environment. This, nevertheless, requires abundant statistical experiments to obtain adequate distributions to model the average behavior of the environment. It is a natural choice to model by nondeterminism a system which is open for interaction from environment, system properties then need to be guaranteed for all potential environments~\cite{Stoelinga02}. Therefore, Markov decision processes (MDPs), which exhibit both nondeterministic and probabilistic behavior, are widely used for modeling reactive systems~\cite{BK2008}. In this paper, we adapted the algorithm for learning deterministic probabilistic finite automata to include nondeterministic actions. Particularly, we learn deterministic labeled Markov decision processes (DLMDPs), where input actions are chosen nondeterministically and outputs given inputs are determined probabilistically, from the observed input and output behavior of a reactive system. This leads to another motivation of the learning purposes. For large systems, we may be interested in only one component, and it receives certain inputs from the environment or other components. Then the learner can output a model which is the representation of this component.

Besides model learning, statistical model checking (SMC)~\cite{Legay10,YS02} techniques can also be used to analyze black-box systems. Statistical model-checking uses hypothesis testing based on sampling runs of a system that allows the user to check to a desired level of confidence whether a given logical property holds with a given (minimum) probability. Unfortunately, this technique is not well suited to MDPs since the presence of nondeterminism making running for sample paths is not well defined~\cite{BFHH11} without an extra scheduler. Moreover, the model output by the model learning approach can be used to other properties without re-sampling, as well as being used for other MDD tasks.

The main contribution of this paper is the development of \ioalergia\ algorithm for learning DLMDP, which is obtained as an adaptation of the previous \alergia~\cite{CarOnc94} algorithm. In order to demonstrate the applicability, the new algorithm is applied to learning models for slot machines from observed system behavior, which is in the form of alternating sequences of inputs and outputs. The evaluation is performed by analyzing and comparing probabilistic linear time properties in the learned model and the known generating model, as well as maximal expected reward and optimal schedulers.

This paper is structured as follows: section~\ref{sec:preliminary} contains background material. Section~\ref{sec:data} describes the procedure of generating learning data, while section~\ref{sec:learning} describes \ioalergia\ algorithm. Section~\ref{sec:experiments} demonstrate its applicability through a case study concerning slot machine. Section~\ref{sec:conclusion} concludes the paper.

\section{Preliminaries}
\label{sec:preliminary}

\subsection{Labeled Markov Decision Processes}
\begin{defn}[LMDP]
A \emph{labeled Markov decision processes (LMDP)} is a tuple ${\mdp}=(Q ,\Sigma_{I},\Sigma_{O},\pi ,\tau , L)$\,

\begin{itemize}
\item $Q $ is a finite set of states,
\item $\Sigma_I$ is a finite input alphabet,  and $\Sigma_O$ is a finite output alphabet,
 \item $\pi :Q \to [0,1]$ is an {initial probability distribution}\/ such that $\sum_{q\in Q }\pi (q)=1$,
\item $\tau :Q  \times \Sigma_{I} \times Q  \to [0,1]$ is the {transition probability function}\/ such that for all $q\in Q $ and all $\alpha \in \Sigma_{I}$, $\sum_{q'\in Q} \tau (q,\alpha,q')=1$, or $\sum_{q'\in Q} \tau (q,\alpha,q')=0$,
\item $L : Q \rightarrow \Sigma_O$\ is a {labeling function}.
\end{itemize}
\end{defn}
An input $\alpha \in \Sigma_I$ is enabled in state $q\in Q $ if and only if $\sum\nolimits_{q' \in Q} {\tau (} q,\alpha ,q') = 1$. Let $Act(q)$ denote the set of enabled actions in $q$.


\begin{defn}[DLMDP]
A LMDP is deterministic if\par
 \begin{itemize}
\item There exists a state $\emph{q}_s \in Q $\ with $\pi (\emph{q}_s)=1$,
\item For all $q\in Q $, $\alpha \in \Sigma_{I}$\ and $\sigma \in \Sigma_{O}$, there exists at most one $q'\in Q $\ with $L (q')=\sigma$\ and $\tau (q,\alpha, q')>0$. We then also write $\tau (q,\alpha,\sigma)$\ instead of $\tau (q,\alpha,q')$.
\end{itemize}
\end{defn}


\subsection{Strings}
Let $\Sigma_O (\Sigma_I \Sigma_O)^{*}$ and $\Sigma_O (\Sigma_I \Sigma_O)^{\omega} $ denote the set of all finite, respectively infinite strings of alternative input and output symbols. For a finite string $s=\sigma_0\alpha_1\sigma_1\ldots\alpha_n\sigma_n$,
$\alpha_i \in \Sigma_I$ and $\sigma_i \in \Sigma_O$, the set of all its prefixes is defined as:
\[
\prefix(s)=\{\sigma_0\alpha_1\sigma_1\ldots\alpha_k\sigma_k\;|\; 0 \leq k\leq n, k \in \Nset \}
\]
For a set of strings $\candS$, $\prefix(\candS)$\ denotes the set of all prefixes of strings $s\in \candS$. We assume an lexicographic ordering on  $\Sigma_O (\Sigma_I \Sigma_O)^{*}$.

In a DLMDP there is a tight connection between strings and states: given an observed string $s$\ there is a unique state $q$\ that the LMDP must be in. Conversely, every state $q$\ is associated with the set $\strings(q)$\ of all strings that lead from the start state to $q$. We therefore use symbols $q$\ for states and $s$\ for strings to some extent interchangeably: $s$\ can also denote the state in a DLMDP reached by the string $s$. The association of strings with states, on the other hand, is not one-to-one. We can still identify $q$\ with the lexicographically minimal $s\in\strings(q)$, and may use $q$\ also to denote this string.

\subsection{Scheduler}

A scheduler~\cite{BK2008} for a MDP \mdp\ is a function $\sch: Q^+ \rightarrow \Sigma_I$ such that $ \sch(q_0q_1 \ldots  q_n)\in Act(q_n)$ for all $q_0,q_1, \ldots, q_n \in Q^+$. The scheduler chooses in any state $q$ one action $\alpha \in \Sigma_I$, and induces a Markov chain, i.e., the behavior of an MDP $\mdp$ under the decisions of scheduler $\sch$ can be formalized by a Markov chain $\mdp_{{\tiny\sch}}$ \cite[Section 10.6]{BK2008}.

%

%
%



A labeled Markov chain (LMC) $\mdp_{\tiny \sch}$ of an LMDP $M$ induced by a scheduler \sch\ defines a probability measure $P_{\tiny \mdp_{\tiny \sch}}$\ on $(\Sigma_O)^{\omega}$\ which is the basis for associating probabilities with events in the LMC $\mdp_{\tiny \sch}$. The probability of a string $s=\sigma_0\sigma_1\ldots\sigma_n, \sigma\in \Sigma_O $\ defined by $\mdp_{\tiny \sch}$ is:  
\[
  P_{{\tiny\mdp_{\sch}}}(s)=\prod_{i=1}^{n} \tau_{\tiny \sch}(\sigma_0 \sigma_1\ldots \sigma_{i-1}, \sigma_i)
\]
where $\tau_{\tiny \sch}$ is the transition probability function of $\mdp_{\tiny \sch}$.




\subsection{Probabilistic LTL}

Linear time temporal logic (LTL) over $\Sigma_O$ is defined as usual by the syntax
\[
\varphi :: = a\;|\; \varphi {}_1 \wedge \varphi _2 \;|\; \neg \varphi \;|\; \bigcirc \varphi \; |\;
\varphi _1  \until \varphi_2 \hspace{5mm} a\in\Sigma_O
\]
For better readability, we also use the derived temporal operators $\Box$\ (always) and $\lozenge$\ (eventually).


Let $\varphi$ be an LTL formula. For $ s= \sigma_0 \sigma_1 \ldots \in  (\Sigma_O) ^\omega$, $s[j\ldots] = \sigma_{j} \sigma_{j+1} \sigma_{j+2} \ldots$\ is the suffix of $s$\ starting with the $(j)$th symbol $\sigma_j$. Then the LTL semantics for infinite words over $\Sigma_O$\ are as follows:\par

\begin{itemize}
\item $s \; \models \; true$
\item $s \; \models \sigma$, iff $\sigma = \sigma_0$
\item $s \; \models \; \varphi_1 \wedge  \varphi_1$, iff $s \; \models \; \varphi_1$ and $ s \; \models \; \varphi_2$
\item $s \; \models \; \neg \; \varphi$, iff $ s \nvDash \varphi $
\item $s \; \models \; \bigcirc \; \varphi$, iff $ s[1\ldots]\models\;\varphi$
\item $s \; \models \; \varphi _1  \until \varphi_2 $, iff $ \exists j\geq 0.\; s[j\ldots]\models \; \varphi_2$ and $s[i\ldots] \models \; \varphi_1$, for all $0 \leq i< j$
\end{itemize}

The syntax of probabilistic LTL (PLTL) is:
\[ \phi :: = P_{\bowtie r} (\varphi)\;\;\; ( \bowtie \; \in \;\geq, \;\leq, \;=;\ r\in[0,1];\ \varphi\in \mbox{LTL}) \]

A labeled Markov decision process $\mdp$ satisfies the PLTL formula $P_{\bowtie r}(\varphi)$ iff $P_{\tiny \mdp_{\tiny \sch}}(\varphi)\,{\bowtie r}$ for all schedulers of $\mdp$, where $P_{\tiny \mdp_{\tiny \sch}}$ is the probability distribution defined by the LMC induced by a scheduler \sch\ of $\mdp$, and $P_{\tiny \mdp_{\tiny \sch}}(\varphi)$ is short for $P_{\tiny \mdp_{\tiny \sch}}(s|s\; \models \;\varphi,s\in (\Sigma_O)^{\omega})$


The quantitative analysis of an MDP $\mdp$ against specification $\varphi$ amounts to establishing the lower and upper bounds that can be guaranteed, when ranging over all schedulers. This corresponds to computing
\[
P^{\pmax}_ {\tiny \mdp}(\varphi ) = \mathop {\sup }\limits_{\tiny \sch} P_{\tiny \mdp_{\sch}} (\varphi )
\;\; \text{and} \;\;
P^{\pmin }_{\tiny \mdp} (\varphi ) = \mathop {\inf }\limits_{\tiny \sch} P_{\tiny \mdp_{\sch}} (\varphi )
\]
where the infimum and the supremum are taken over all schedulers for $\mdp$.
\section{Data Generation}
\label{sec:data}

The data we learn from is generated by observing the running reactive system. From the system we can observe input actions which determine probability distributions over successor states, and outputs which are labels of successor states. The learning algorithm requires that all nondeterministic choices are resolved by a \emph{fair} scheduler \sch\ which means each input action will be chosen infinitely often. We assume that the input and output will be observed alternately, and every observation sequence starts from the label of the initial state, and ends in a state, i.e. $\sigma_0\alpha_1\sigma_1\ldots\alpha_{n}\sigma_{n}, \text{with} \; \alpha_i \in \Sigma_I \;\text{and}\; \sigma_i \in \Sigma_O $.


Usually, enabled and disabled actions for states in a black-box system are unknown. Therefore, we allow that all actions can be chosen on each state of the system. For enabled actions, the system will transit to other states, and the input and the corresponding label of the successor state will be collected. For disabled actions, the system will stay in the same state but give a special error message. Through this setting, enabled and disabled inputs could be distinguished. Furthermore, we denote the prompted error by $err$, thus the output alphabet $\Sigma_{O}$ is extended to $\Sigma_{O} \cup \{err\}$. Due to the memoryless scheduler, the same disabled input on the same state could be chosen more than once, and the statistic information about $err$ will be found necessary in the following \emph{compatibility} test.


%
After all nondeterministic choices have been resolved, let $S_1^{\omega}, S_2^{\omega},\ldots$\ be an independent family of $P_{\tiny\mdp_{\tiny \sch}}$-distributed random variables (with values in $\Sigma_O (\Sigma_I \Sigma_O)^{\omega}$), and $L_1,L_2,\ldots$\ be an independent family of integer-valued random variables, such that the $L_i$\ are also independent of the $S_i^{\omega}$. We assume that we observe the finite observation sequences $S_i:= \sigma_0\alpha_1\sigma_1\ldots\alpha_{L_{i }}\sigma_{L_{i}}$, i.e., the first $L_i$\ symbols of $S_i^{\omega}$. Thus, we observe the independent run of the system for a period of time that is determined independently of the observed behavior (in particular, the observation does not automatically end when the system enters a deadlock or failure state -- such a situation would rather lead to repeated deadlock or failure observations in the final part of the sequence). We assume that the $L_i$\ are unbounded, i.e. $P(L_i>k)>0$\ for all $k \in \Nset$. This will be satisfied by a geometric distribution for the $L_i$. For some models, there exists a uniquely labeled absorbing state which can be identified by its observation (e.g., a failure state which can not recover). When prior knowledge is available, observations can be stopped when the model reaches that state.

Finally, we denote with $\candS[n]=S_1,\ldots,S_n$\ the sample consisting of the first $n$\ observations.




\section{Learning}
\label{sec:learning}

\ioalergia\ for learning DLMDP consists of two phases. Firstly, represent the data as I/O frequency prefix tree acceptor (\IOFPTA) where common prefixes are combined together. Then, do \emph{compatibility} test on the tree following lexicographical order. If two states are compatible which requires that the next state distributions given the same input are compatible, they and their successor states will be merged correspondingly.

\subsection{\IOFPTA}

The \emph{input and output frequency prefix tree acceptor} \IOFPTA\ is constructed as a representation of the set of strings $\candS$ which captures the behavior of the reactive system under observation. Since in DLMDP, same sequences will lead to the same state, then in \IOFPTA\ common prefixes are merged together and result in a tree shaped automaton. Each node in the tree is labeled by an output symbol $\sigma \in \Sigma_O$, and each edge is labeled by an input action $\alpha \in \Sigma_I$. Every path from the root to a node corresponds to a string $s\in \prefix(\candS)$. The node $s$\ is associated with the frequency function $f(s,\alpha,\sigma)$\ $(\alpha\in\Sigma_I,\; \sigma \in \Sigma_O)$\, which is the number of strings in $\candS$\ with the prefix $s\alpha\sigma$, and $f(s,\alpha) = \sum \nolimits_{\sigma \in \Sigma_O} {f(s,\alpha,\sigma)}$. From one node in \IOFPTA, given an input action and an output symbol, the next state can be uniquely determined. An \IOFPTA\ can be transformed to DLMDP by normalizing frequencies $f(s,\alpha,\cdot)$\ to $\tau(s,\alpha,\cdot)$. As assumed in data generation phase, when the scheduler chooses a disabled input on a state in LMDP, the model will stay in the current state, and output the symbol $err$. We are going to take the special meaning of the $err$ symbol into account in the \IOFPTA\ construction. Specifically, $s$ and $s\alpha err$ would lead to the same state from the root state. We will take the special treatment for the err symbol, but there is no difference between it and other symbols in learning. A new node labeled by $err$\ will not be created as a successor node or we can say that the $err$ nodes are folded up.

\begin{example}{\IOFPTA} \label{exp:tree}

\begin{figure}[htbp]
 \centering
 \includegraphics[scale=1]{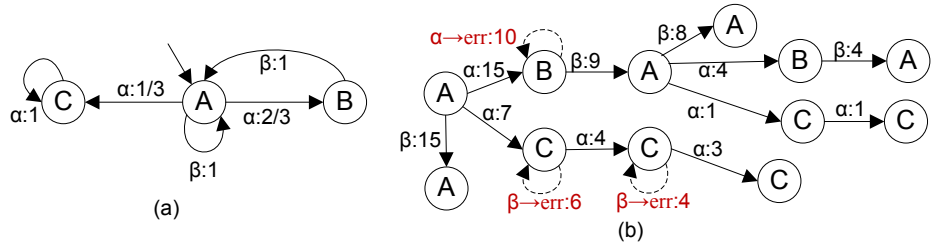}
 \caption{(a) A DLMDP over $\Sigma_O=\{A,B,C,err\}$ and $\Sigma_I=\{\alpha,\beta\}$; (b) The corresponding \IOFPTA.}
  \label{fig:IOFPTA}
 \end{figure}

The \IOFPTA\ in Fig.\ref{fig:IOFPTA}(b) is constructed from sample sequences generated by a DLMDP $\mdp$\ in Fig.\ref{fig:IOFPTA}(a). The root node is labeled by $A$. From the root, given input $\alpha$, successor nodes which are labeled by $B$\ and $C$, will be reached by strings with the prefix $A\alpha B$\ or $A\alpha C$, respectively. For the frequency, $f(A,\alpha,B)=15$\ and $f(A,\alpha,C)=7$. The input action $\beta$ is disabled in the state with label $C$ of (a).  Then the tree will stay in node labeled by C when we meet the input $\beta$ (which is drawn and linked by dash lines in Fig.\ref{fig:IOFPTA}(b)). Then for each node the incoming frequencies are not equivalent to the outgoing frequencies.

\end{example}

\subsection{\ioalergia}

\ioalergia\ algorithm, is an adapted version of the \alergia\ algorithm \cite{CarOnc94,Higuera10}. As seen in Example~\ref{exp:tree}, the same state in generating LMDP could be reached by more than one sequences through running, which will create more than one node in the \IOFPTA. The basic idea of this learning algorithm is to approximate the generating model by grouping together the nodes in \IOFPTA\ which can be mapped to the same state in the generating model. The partition which is introduced by grouping nodes will be inferred by pairwise testing. The \emph{compatibility} of two nodes is tested by comparing distributions defined by nondeterministic choices, and recursively testing on successor nodes. If two nodes in the tree pass the \emph{compatibility} test which means they can be mapped to the same state in the generating model, then they will be merged, as well as their successor nodes.

\newcommand{\merged}{\emph{merged}}
\newcommand{\false}{\emph{false}}
\newcommand{\true}{\emph{true}}

\begin{algorithm}[!htbp]
\caption{\ioalergia}
    \label{algo:LMDP-learner}
    \begin{algorithmic}[1]
    \REQUIRE: A dataset $\candS$ and a parameter $\epsilon \in (0,1]$;
    \ENSURE: A DLMDP $A$;
    \STATE $T, A \leftarrow \text{IOFPTA}(S)$; \label{step:initfpta} \\
    \STATE $\RED \leftarrow \emph{q}_s^A $; \\
    \STATE $\BLUE \leftarrow \{q \mid q=\emph{q}_s^A\alpha\sigma, \alpha \in \Sigma_I,\sigma \in \Sigma_O, \emph{q}_s^A\alpha\sigma \in \prefix(\candS)\}$;  /* immediate successor states  */ \\
    \WHILE { $\BLUE \neq \emptyset$  \label{step:LMC-learner_loop1} }
         \STATE $q_b \leftarrow$\ lexicographically minimal $q\in\BLUE$; \\
         \STATE $\merged \leftarrow \false$; \\   
         \FOR { $q_r\in\RED\ $  /* in lexicographic order */  \label{step:LMC-learner_loop2_begin} }
              \IF {$\COMPATIBLE(T,  q_r , q_b ,\epsilon)$ \label{step:comptest} }
                    \STATE $A \leftarrow \MERGE(A,q_r,q_b)$; \label{step:merge}\\
                    \STATE $\merged \leftarrow \true$; \\
              \ENDIF
         \ENDFOR \label{step:LMC-learner_loop2_end} \\
        \IF {$!\merged$ }
             \STATE $ \RED \leftarrow \RED \cup \{{q_b} \}$; \\
        \ENDIF

        \STATE {$\BLUE  \leftarrow \BLUE\setminus \{q_b\} \cup \{ q=q_r \alpha\sigma \mid \alpha\in\Sigma_I,\sigma\in\Sigma_O, q \in \prefix(\candS), q_r \in \RED,  q \notin \RED \}$ }; \\

    \ENDWHILE \\
    \RETURN\ $\emph{makeDLMDP}(A)$; /* normalize */ \\ \label{step:LMC-learner_normalize}
    \end{algorithmic}
\end{algorithm}

In the learning algorithm, firstly, two \IOFPTA s $T$\ and $A$\ are constructed as the representation of the dataset $\candS$\ (line~\ref{step:initfpta} of the Algorithm~\ref{algo:LMDP-learner}). The \IOFPTA\ $T$\ is kept as a data representation from which relevant statistics are retrieved during the execution of the algorithm. The \IOFPTA\ $A$\ is iteratively transformed by merging nodes which have passed the \emph{compatibility} test. All compatibility is tested on $T$, and the reason for this is that it has a clear interpretation as empirical probabilities defined by the data $\candS$. Following the terminology from \cite{Higuera10}, Algorithm~\ref{algo:LMDP-learner} maintains two sets of states: \RED states, which have already been determined as representative states of partitions and will be included in the final output DLMDP, and \BLUE states which are going to be tested. Initially, \RED contains only the initial state while \BLUE contains the immediate successor states of the initial state. During iterations, the lexicographically minimal node $q_b$\ in \BLUE\ will be chosen. If there exists a state $q_r$\ in \RED\ which is compatible with $q_b$, then $q_b$\ and its successor nodes are going to be merged into $q_r$ and its corresponding successor states. If $q_b$\ is not compatible with any state in \RED, it will be included in \RED. At the end of each iteration, \BLUE\ is going to be updated as the margin between \RED\ and the remaining states, in the other word, the set of states which are immediate successor states of \RED\ but not included in it. After merging all possible compatible nodes in the tree, the frequencies in $A$\ are going to be normalized by the Algorithm~\ref{algo:LMDP-learner} (line \ref{step:LMC-learner_normalize}). Then a DLMDP is constructed.


\subsection{Compatibility Test}

Algorithm~\ref{algo:comp} demonstrates the \emph{compatibility} test. It will return true if two nodes are compatible, i.e., the distance of distributions for every action is within the Hoeffding bound~\cite{Hoeffding63}, Algorithm~\ref{algo:hoeffding}, parameterized by $\epsilon$. Formally, two nodes $q_r$ and $q_b$ are $\epsilon$\emph{-compatible} ($1\geq\epsilon>0$), if it holds that:
\begin{enumerate}
\item $L(q_r)=L(q_b)$
\item $\text{\emph{Hoeffding}}(f(q_r,\alpha,\sigma),
    f(q_r,\alpha),f(q_b,\alpha,\sigma),f(q_b,\alpha),\epsilon)$ is TRUE, for all $\alpha \in \Sigma_I$ and $\sigma\in \Sigma_O$ .
\item Nodes $ q_r\alpha\sigma $ and $ q_b \alpha \sigma$ are $\epsilon$\emph{-compatible}, for all $\alpha \in \Sigma_I$, and $\sigma\in \Sigma_O$
\end{enumerate}
Condition 1) requires two nodes in the tree to have the same label. Condition 2) defines the compatibility between each outgoing transition with the same input action respectively from state $q_r$ and $q_b$. The last condition requires the compatibility to be recursively satisfied for every pair of successors of $q_r$ and $q_b$. If two nodes in \IOFPTA\ are compatible, then distributions for all input actions should pass the \emph{compatibility} test.

In the original \alergia\ algorithm, termination probabilities of two nodes are compared, while not in  Algorithm~\ref{algo:comp}. The reason is that the termination probability is not included in the definition of DLMDP. In Algorithm~\ref{algo:hoeffding}, the distance of two empirical probabilities are compared with the \emph{Hoeffding} bound. If there is few, even none, statistical evidence to support their difference, the distance is small. In particular, two nodes are compatible, if there is no evidence against that. The $err$ information is used to discriminate two nodes which have different enabled actions. For example, there are $q_1$ and $q_2$, and input action $\alpha$ is only enabled on $q_1$. For $q_1$, $f(q_1,\alpha,\sigma)>0, \sigma \neq err$ and $f(q_1,\alpha,err)=0$, while $f(q_2,\alpha) = f(q_2,\alpha, err)> 0$. Comparing the empirical probability distribution over $\Sigma_O$ including $err$, $q_1$ and $q_2$ can not be compatible.


\begin{algorithm}[h]
    \caption{\COMPATIBLE}
    \label{algo:comp}
    \begin{algorithmic}[1]
    \REQUIRE: IOFPTA $T$ , nodes $q_r$ and $ q_b$, $\epsilon \in (0,1]$
    \ENSURE: \emph{true} if $q_r$\ and $q_b$\ are compatible
    \IF {$L(q_r) \neq L(q_b) $\ }
    \RETURN\ \false
    \ENDIF

    \FOR{ $\alpha \in\Sigma_I$}
        \FOR{ $\sigma\in\Sigma_O$}
           \IF{$!\text{\emph{Hoeffding}}(f^T(q_r,\alpha,\sigma),f^T(q_r,\alpha) f^T(q_b,\alpha,\sigma),f^T(q_b,\alpha),\epsilon)$ }
                \RETURN\  \false
           \ENDIF
            \IF {$ !\COMPATIBLE(T, q_r\alpha\sigma, q_b\alpha\sigma,\epsilon$) }
                \RETURN\  \false
           \ENDIF
        \ENDFOR
     \ENDFOR
     \RETURN\ \true

    \end{algorithmic}
\end{algorithm}

\begin{algorithm}[h]
    \caption{\emph{Hoeffding}}
    \label{algo:hoeffding}
    \begin{algorithmic}[1]
    \REQUIRE: $f_1,n_1, f_2, n_2, \epsilon \in (0,1]$
    \ENSURE: \emph{true} if $f_1/n_1$ and $f_2/n_2$ are sufficiently close
     \IF {$n_1 ==0$\ or $n_2 ==0$ }  \label{step:Hoeffding_no}
     \label{step:zerotest}
           \RETURN \ \true
     \ENDIF

     \RETURN\ $|\frac{{f_1}}{{n_1}} - \frac{{f_2}}{{n_2}}| < (\sqrt {\frac{1}{{n_1}}}  + \sqrt {\frac{1}{{n_2}}} ) \cdot \sqrt {\frac{1}{2}\ln \frac{2}{\epsilon }}$

    \end{algorithmic}
\end{algorithm}

\subsection{Merge states}

If two states $q_r$ and $q_b$ are \emph{compatible}, $q_b$ will be merged to $q_r$. The \MERGE\ procedure (line \ref{step:merge} of the Algorithm~\ref{algo:LMDP-learner}) follows the same way as described in \cite{Higuera10}: firstly, the (unique) transition leading to $q_b$\ from its predecessor node $q'$ ($f^A(q',\alpha, q_b)>0$) is re-directed to $q_r$ by setting $f^A(q',\alpha,q_r) \leftarrow f^A(q',\alpha,q_b)$\ and $f^A(q',\alpha,q_b)=0$. Then, successor nodes of $q_b$ will be recursively folded to the corresponding successor nodes of $q_r$.


\begin{example}{Merge States}
\begin{figure}[!htbp]
 \centering
 \includegraphics[scale=1]{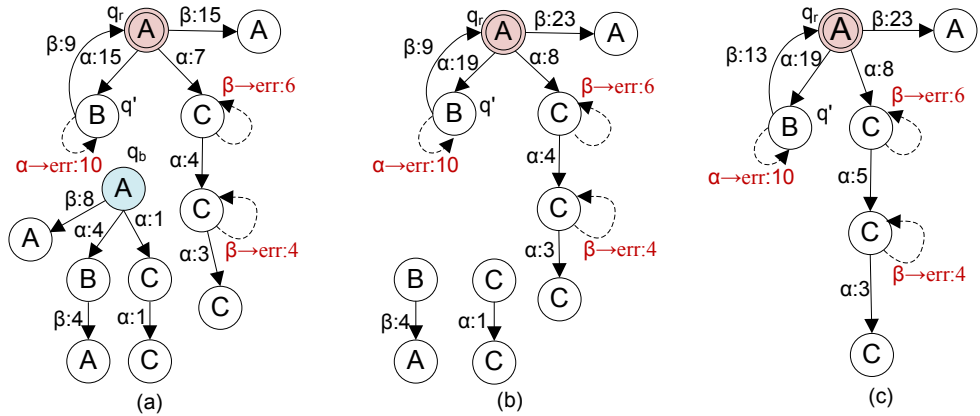}
\caption{Merge states $q_r$ and $q_b$}
  \label{fig:merge}
 \end{figure}


Fig.~\ref{fig:merge} shows the procedure that the node $q_b$ (shadowed) will be merge to the node $q_r$ (shadowed double circle). In (a), the transition from the node $q'$ to $q_b$ firstly redirected to $q_r$. In (b), transitions from $q_b$ to three successor nodes labeled with A, B and C, will be folded into the corresponding successor nodes of $q_r$, respectively. (c) illustrates the result after merge.

\end{example}


\subsection{Discussion}
The algorithm takes a set $\candS$ of sample sequences and a parameter $\epsilon$\ as inputs. Here $\epsilon$\ is used to bound the type-\uppercase\expandafter{\romannumeral1} error, which is the probability of wrongly rejecting a correct compatibility hypothesis. Smaller values of $\epsilon$\ lead to loose Hoeffding bounds and making \ioalergia\ output a smaller model. For any particular finite sample size we try to tune the choice of $\epsilon$\ so as to obtain the best approximation to the real model. In order to do this we run \ioalergia\ with different $\epsilon$\ values, and evaluate the learned model using the \emph{Bayesian Information Criterion (BIC)} score. This score combines the likelihood of a model with a term penalizing model complexity. Concretely, the BIC score of a DLMDP $A$\ given data $\candS$\ is defined as
\begin{displaymath}
  \emph{BIC}(A\mid \candS):=
  \emph{log}(P_A(\candS))-1/2\sizeof{A}\emph{log}(N)
\end{displaymath}
where $\sizeof{A}=\sizeof{Q} \cdot \sizeof{\Sigma_I} \cdot \sizeof{\Sigma_O}$ is the number of free parameters in the model. $N$\ is the number symbols in the data. Using a golden section search~\cite[Section E.1.1]{TanSteKum06} we systematically search for an $\epsilon$\ value maximizing the BIC score of the learned model. Our algorithm is implemented in Matlab and is available for download at \url{http://mi.cs.aau.dk/code/ioalergia}.

A convergence analysis, similar to the analysis in~\cite{MCJN+11,HigTho00} for deterministic Markov chain models, can be obtained for \ioalergia: first, one can show that in the large sample limit, \ioalergia\ will identify up to bisimulation equivalence the structure of the true model from which the data was sampled; the structure of a model refers to all of its components, except the probability values of transitions. Second, the parameters in the learned model will converge to the corresponding parameter values in the true model. As a slight refinement of Theorem 2 in~\cite{MCJN+11}, one then obtains that for any LTL formula $\varphi$:
\[
P( \lim_{n\rightarrow\infty}P^{\pmax}_{A^n}(\varphi)  = P^{\pmax}_{\tiny \mdp}(\varphi)) =1, \text{and}  \hspace{10mm} P( \lim_{n\rightarrow\infty}P^{\pmin}_{A^n}(\varphi)  = P^{\pmin}_{\tiny \mdp}(\varphi)) =1;
\]
where $A^n$\ is the DLMDP returned by \ioalergia\ on data $\candS[n]$. As also observed in~\cite{MCJN+11}, similar results do not carry over to PCTL formulas.

\section{Experiments}
\label{sec:experiments}
In this section, we are going to show the applicability of the \ioalergia\ algorithm using a case study based on the slot machine~\cite{Jansen02b}. The slot machine we considered has 3 reels, named as \emph{reel}-1, \emph{reel}-2 and \emph{reel}-3, and each reel contains 5 different symbols: \emph{lemon}, \emph{grape}, \emph{cherry}, \emph{bar} and, \emph{apple}. The slot machine will return a prize based on the combination of symbols on those 3 reels. The prizes for different configurations are shown in Table~\ref{tab:prize}(a). We extend the basic gambling machine as follows: at each round the player can choose one of the reels to spin, and other reels will be kept. The player starts with paying 1 coin for first 3 spins, and afterwards each extra spin costs 1 additional coin. Each reel must be spun at least once, and the player can quit the game only if all reels have been spun. The behavior of the slot machine contains both probabilistic and nondeterministic aspects. Specifically, the symbol show for each reel is probabilistic, but the choice of which reel to spin is nondeterministic.

%


%

 \begin{table}[htbp]

 \begin{minipage}[b]{0.4\textwidth}
  \centering
 \caption{Prize}
\begin{tabular}{|c|c|c|c|}

\hline
 \emph{reel}-1 &  \emph{reel}-2 & \emph{reel}-3 & Prize\\
\hline
bar & bar & bar & 10\\
\hline
cherry & cherry & cherry & 10\\
\hline
grapes & grapes & grapes & 10\\
\hline
? & bar & bar & 5\\
\hline
cherry & ? & cherry & 5\\
\hline
grapes & grapes & ? & 5\\
\hline
? & ? & bar & 2\\
\hline
? & ? & cherry & 1\\
\hline
\end{tabular}
\label{tab:prize}
\end{minipage}
 \hspace{0.5cm}
\begin{minipage}[b]{0.6\textwidth}
\centering
\caption{Summery of slot machines}
  \label{tab:model}
\setlength{\extrarowheight}{2pt}
  \centering
  \setlength{\tabcolsep}{4pt}
  \begin{tabular}{|c|l r ||l r |}\hline
           & \multicolumn{2}{c||}{ \tabincell{c}{Deterministic \\ Slot Machine} } & \multicolumn{2}{c|}{\tabincell{c}{Nondeterministic \\ Slot Machine}} \\ \hline
 N&   $|Q|$ & $|\tran|$ & $|Q|$ & $|\tran|$  \\ \hline
4 & 437&4021&510&4959\\
\hline
6 & 867&10721&1012&13291\\
\hline
8 & 1297&17421&1514&21623\\
\hline
10 & 1727&24121&2016&29955\\
  \hline

  \end{tabular}
\end{minipage}
\end{table}

In the following parts of this section, the algorithm will be applied for learning deterministic and nondeterministic models for different number of spins. A memoryless and random scheduler with a uniform distribution over all input actions, that modeling the \emph{fair} requirement, is used in the data generation procedure. For experiment, we analyze the behavior of learned models by comparing them with known generating models in terms of maximal and minimal probabilities of winning a specific reward as well as the maximal expected reward in general. These probabilities and rewards are all computed by PRISM~\cite{KNP11}. We will also analyze the accuracy that the optimal action in the learned model given symbols on reels and number of times the reels have been spun.

\subsection{Learning models from Deterministic systems}
We implemented the slot machine in PRISM. The distribution for 3 reels showing different symbols are $(0.2,0.2,0.1,0.3,0.2)$, $(0.2,0.1,0.3,0.2,0.2)$, and $(0.2,0.3,0.2,0.1,0.2)$, respectively. In this model, there are 4 actions: spin \emph{reel}-1 ($sp_1$), spin  \emph{reel}-2 ($sp_2$), spin \emph{reel}-3 ($sp_3$), and get the prize \emph{(pay)}, thus $\Sigma_I=\{sp_1,sp_2,sp_3, pay\}$. Every state is labeled by the combination of states on the 3 reels and the number of times the reels have been spun. We also attached reward variables to the states which are labeled by \emph{prize}. Table~\ref{tab:model} shows statistics for models with various number of spins. Here, $N$ ($N\geq3$) is the number of spins, $|Q|$ is the number of states, and $|\tran|$ is the number of transitions. \\

The generating model is a deterministic LMDP. The results of applying the learning algorithm for different data sets are produced by the generating model are summarized in Table~\ref{tab:exp1}: $|\candS|$ is the number of symbols in the dataset ($\times 10^3$), $|\seq|$ is the number of sequences in the dataset; $|\text{IOFPTA}|$ is the number of nodes in the IOFPTA; Time is the learning time (in seconds), including the time for constructing \IOFPTA\ and the average time for each iteration performed by the golden section search (typically the golden section search terminated after 14 to 19 iterations); `$\epsilon$ range' is the interval (identified using the golden section search) for $\epsilon$ for which a BIC-optimal DLMDP is learned, $|Q|$ is the number of states in the learned model.

\begin{table}[htbp]
  \caption{Experimental results for the slot machine models.}
\setlength{\extrarowheight}{2pt}
  \centering
  \setlength{\tabcolsep}{4pt}
  \begin{tabular}{|cc|c|c|c|c|c |}\hline
&$|\candS|(\times 10^3)$ &  $|\seq|$  &  $|\text{IOFPTA}|$  & Time & $\epsilon \text{ range}$ & $|Q|$    \\ \hline
\multirow{4}{0.2cm}{
  \begin{turn}{90} $N=4$   \end{turn} }

 & 160 & 5832 & 20915 & 9.8 & [0.0020; 0.1552] & 436\\
 & 640 & 23246 & 48373 & 29.9 & [0.0020; 0.1552] & 437\\
 & 1280 & 46374 & 64064 & 50.2 & [0.0020; 0.1250] & 437\\
\hline

\multirow{4}{0.2cm}{
  \begin{turn}{90} $N=6$   \end{turn} }

  & 160 & 5779 & 33829 & 16.0 & [0.0020; 0.1553] & 866\\
 & 640 & 23154 & 122458 & 46.9 & [0.0020; 0.1553] & 867\\
 & 1280 & 46273 & 231029 & 84.9 & [0.0010; 0.0776] & 867\\

\hline

\multirow{4}{0.2cm}{
  \begin{turn}{90}  $N=8$   \end{turn} }

 & 640 & 23054 & 148225 & 66.1 & [0.0020; 0.1553] & 1297\\
 & 1280 & 46242 & 283749 & 116.6 & [0.0010; 0.0776] & 1297\\
 & 2000 & 72284 & 429555 & 153.0 & [0.0010; 0.0776] & 1297\\

\hline

\multirow{4}{0.2cm}{
  \begin{turn}{90} $N=10$   \end{turn} }
 & 1280 & 46241 & 317794 & 142.5 & [0.0005; 0.0388] & 1725\\
 & 2000 & 72250 & 482943 & 184.0 & [0.0005; 0.0313] & 1727\\
 & 5000 & 180755 & 1135055 & 454.4 & [0.00006; 0.0040] & 1727\\
\hline
\end{tabular}
\label{tab:exp1}
\end{table}


\begin{figure}[!htb]
 \centering
 \includegraphics[scale=0.8]{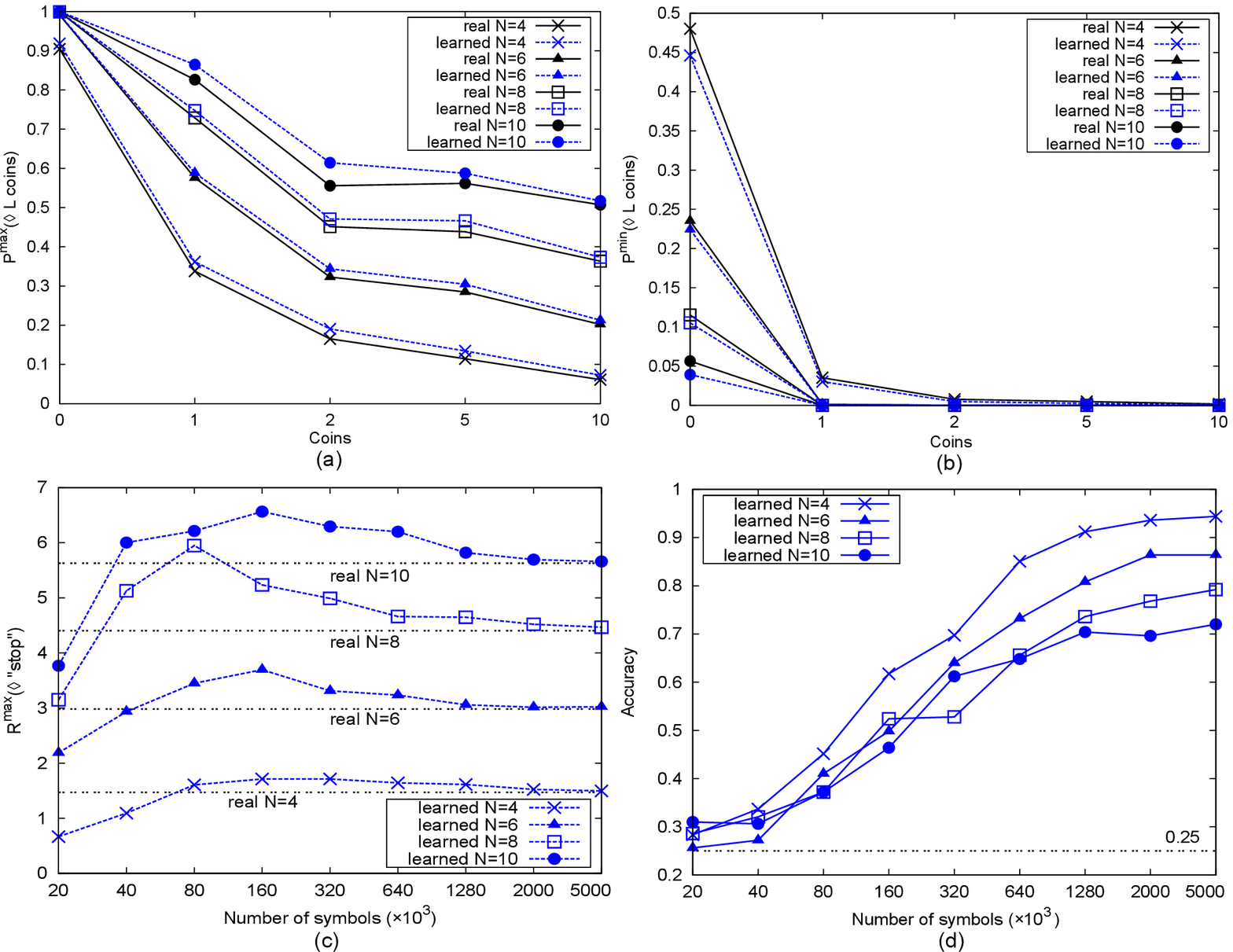}
\footnotesize{\caption{Evaluation results for learning deterministic models. Fig (a) and (b): The maximal and minimal probabilities of eventually being awarded L coins given 4, 6, 8, and 10 initial spins, here L$\in \{0,1,2,5,10\}$. As shown, the model for $N=4$ is learned from $1280\times 10^3$ symbols, and models for $N=6,8,10$ are all learned from $5000\times 10^3$ symbols. Fig (c), shows maximal rewards ($R^{\pmax}(\lozenge\; stop)$) in learned models and the generating model. In Fig (d), the accuracy of the optimal action the learned models is shown.}\label{fig:deter}}

\vspace{-0pt}
\end{figure}

Fig.~\ref{fig:deter}(a) and (b) show the maximal and minimal probabilities of eventually getting different prizes using the $P^{\pmax}(\lozenge \text{L}\; coins)$ and $P^{\pmin}(\lozenge \text{L}\; coins)$, where L$\in\{0,1,2,5,10\}$ (on both generating model and learned models, for $N=4,6,8,10$). As the size of dataset increases, the learned models provide better approximations of the maximal and minimal probabilities defined for the generating models. Using PRISM, the maximal expected reward for one gamble ($R^{\pmax}(\lozenge\; stop)$) can be computed. In Fig.~\ref{fig:deter}(c), for various initially bought spin chances, the maximal expected rewards for the learned models (dashed lines) are all approaching the ones for the generating models as the sizes of the datasets increase.

The optimal action which reel to spin next for a specific configuration of the reels, can also be accurately preserved by learned models. For example, given that there are three \emph{apples} on reels and we only have 1 spin left, the best choice is to spin the 3rd reel since taking any other action will not produce a prize. We consider the 125 configurations where every reel has been spun once.  Given a specific configuration $C_i$, the optimal action in the learned model and the generating model are denoted as $Act_i^l$ and $Act_i^g$, respectively. We define a criterion which interpret the accuracy of optimal actions inferred by the learned model against the generating model as follows:
\[Acc = \sum\nolimits_{i = 1}^{125} {P^{\pmax}(C_i)  \cdot \frac{{|Act_i^l \cap Act_i^g |}}{{|Act_i^l |}}}\]
Where, $P^{\pmax}(C_i)$ is the maximal probability of reaching configuration $C_i$. As shown in Fig.~\ref{fig:deter} (d), by increasing the size of dataset, the learned models have almost the same optimal actions as the generating models. Even with very limited data amount, accuracies for optimal actions in learned models are always greater than 25\%, which is the probability of randomly choosing an optimal action.


\subsection{Learning models from Nondeterministic systems}

In order to make the slot machine more interesting, we increase the prize for three \emph{bar}s but reduce the probability of getting that. This is done by adding another \emph{bar} on \emph{reel}-2, two \emph{bar}s, denoted as $b_1$ and $b_2$, that are indistinguishable, but have different mechanical characteristics. The probability for these two bars depend on the symbols on other two reels.

The distributions for all reels are shown in Table~\ref{tab:reel13} and Table~\ref{tab:reel2}. Since reels are no longer independent, we name refer to machine as \emph{hooked slot machine}. In this machine, the probability of getting 3 \emph{bar}s is decreased, but the reward for getting 3 bars is 20 coins. Every other configuration has the same prize as the previous game. After this modification, the generating model becomes nondeterministic, and its statistics listed in Table~\ref{tab:model}.


\begin{table}[htbp]
\caption{Probability distributions for 3 reels}
  \setlength{\tabcolsep}{4pt}
\centering
\subtable[Probability distributions for the 1st and the 3rd reel]{
  \begin{tabular}{|cc|c|c|c|c|c |}\hline
&  &   lemon   &   grape   & cherry & bar & apple    \\ \hline
 \multirow{4}{0.2cm}{
  \begin{turn}{90} $reel1$   \end{turn}  }

&$r_2=b_1$ & 0.2 & 0.2 & 0.1 & 0.3 & 0.2\\
&$r_2=b_2$ & 0.3 & 0.2 & 0.1 & 0.05 & 0.35\\
&other & 0.25 & 0.2 & 0.1 & 0.15 & 0.3\\
\hline

\multirow{4}{0.2cm}{
  \begin{turn}{90}  $reel3$   \end{turn} }
&$r_2=b_1$ & 0.2 & 0.3 & 0.2 & 0.05 & 0.25\\
&$r_2=b_2$ & 0.1 & 0.3 & 0.2 & 0.3 & 0.1\\
&other & 0.2 & 0.3 & 0.2 & 0.15 & 0.15\\
\hline
\end{tabular}
  \label{tab:reel13}
}
\qquad
\subtable[Probability distributions for 2nd reel]{
\begin{tabular}{|c|c|c|c|c|}
\hline
 & $r_1=b$ & $r_3=b$ & $r_1,r_3=b$ & other\\
\hline
lemon & 0.2 & 0.2 & 0.26 & 0.2\\
\hline
grape & 0.1 & 0.1 & 0.1 & 0.1\\
\hline
cherry & 0.3 & 0.3 & 0.3 & 0.3\\
\hline
bar 1 & 0.18 & 0.02 & 0.02 & 0.1\\
\hline
bar 2 & 0.02 & 0.18 & 0.02 & 0.1\\
\hline
apple & 0.2 & 0.2 & 0.3 & 0.2\\
\hline
\end{tabular}
  \label{tab:reel2}
}
\end{table}

In this experiment, we apply \ioalergia\ for learning DLMDPs from data generated by the nondeterministic models. The learning results are summarized in Table~\ref{tab:exp2}, where each column has the same meaning as in Table~\ref{tab:exp1}. Given sufficient dtat, we observed that learned models have the same number of states as the deterministic models of the previous slot machine, thus the states introduced by the extra symbol on \emph{reel}-2 was not get identified. The reason is that states labeled by \emph{$b_1$ on reel-2} and \emph{$b_2$ on reel-2} are mixed and generally observed as \emph{bar on reel-2}.




\begin{table}[htbp]
  \caption{Experimental results for hooked slot machines.}
\setlength{\extrarowheight}{2pt}
  \centering
  \setlength{\tabcolsep}{4pt}
  \begin{tabular}{|cc|c|c|c|c|c |}\hline

&$|\candS|(\times 10^3)$ &  $|\seq|$  &  $|\tree|$  & time & $\epsilon \text{ range}$ & $|Q|$\\
 \hline
\multirow{4}{0.2cm}{
  \begin{turn}{90}   $N=4$  \end{turn} }

&160 & 5794 & 20768 & 9.7 & [0.0020; 0.1552] & 437\\
&640 & 23185 & 48530 & 29.8 & [0.0020; 0.1250] & 437\\
&1280 & 46308 & 64354 & 51.5 & [0.0010; 0.0776] & 437\\
\hline

\multirow{4}{0.2cm}{
  \begin{turn}{90}   $N=6$  \end{turn} }

&160 & 5737 & 33755 & 15.7 & [0.0039; 0.2500] & 867\\
&640 & 23174 & 122575 & 46.6 & [0.0020; 0.1552] & 867\\
&1280 & 46380 & 231260 & 84.0 & [0.0005; 0.0388] & 867\\

\hline

\multirow{4}{0.2cm}{
  \begin{turn}{90}  $N=8$   \end{turn} }

&640 & 23143 & 148730 & 63.7 & [0.0020; 0.1552] & 1297\\
&1280 & 46260 & 284310 & 112.7 & [0.0010; 0.0776] & 1297\\
&2000 & 72212 & 430102 & 166.1 & [0.0005; 0.0313] & 1297\\
\hline

\multirow{4}{0.2cm}{
  \begin{turn}{90}   $N=10$  \end{turn}  }
&1280 & 46371 & 318423 & 138.6 & [0.0010; 0.0776] & 1723\\
&2000 & 72360 & 483696 & 202.5 & [0.0005; 0.0313] & 1724\\
&5000 & 180781 & 1135149 & 460.3 & [0.0010; 0.0625] & 1725\\
\hline
\end{tabular}
\label{tab:exp2}
\vspace{0pt}
\end{table}

\begin{figure}[!htb]
 \centering
 \includegraphics[scale=0.8]{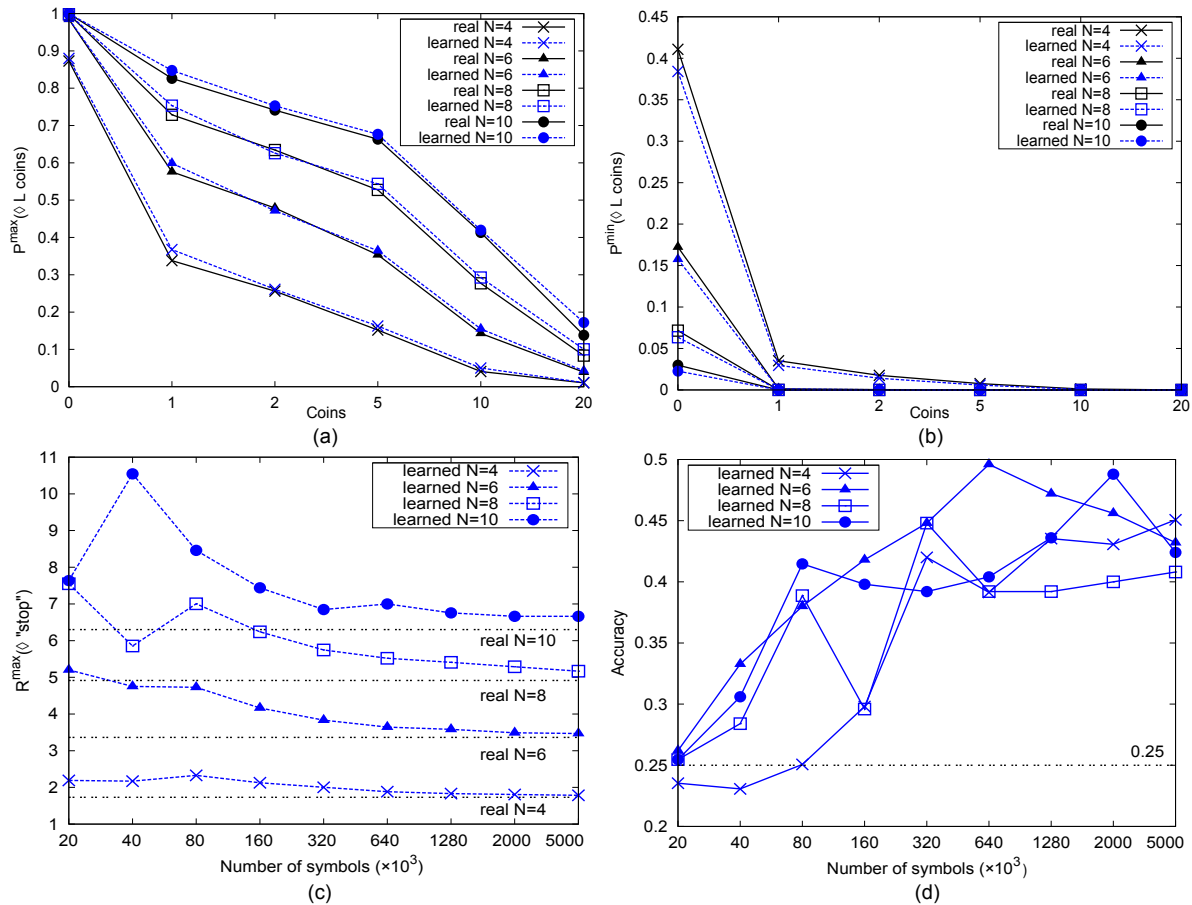}
\caption{Evaluation results for learning nondeterministic models. (a) and (b) are the maximal and minimal probability of eventually being awarded $L$ coins  $L \in \{0,1,2,5,10\}$.  The size of each dataset is the same as  Fig.~\ref{fig:deter}. (c): maximal rewards computed by $R^{\pmax}(\lozenge\; stop)$ in learned models and generating models. (d): the accuracy of optimal actions suggested by learned models.}
\label{fig:non}
\vspace{-0pt}
\end{figure}

Fig.~\ref{fig:non} shows maximal and minimal probabilities for getting different prizes, maximal rewards from the initial state and the accuracy of the optimal action. Given adequate data, learned deterministic models provide good approximations for nondeterministic generating models in terms of maximal probability, minimal probability and the maximal expected reward. On the other hand, the accuracy of choosing optimal action in next step is no longer as good as before. Nevertheless, the suggestion given by learned model is still better than random choice (which has 25\% accuracy) in most cases.

The generating model is a nondeterministic LMDP, so there is no guarantee that the learned model preserves all PLTL properties . For example, suppose there are two \emph{bar}s after two spins, and corresponding to the configurations `\emph{bar}, \emph{bar}, \emph{not-spun}' ($C_1$), `\emph{bar}, \emph{not-spun}, \emph{bar}' ($C_2$), and `\emph{not-spun}, \emph{bar}, \emph{bar}'\ ($C_3$). From these configurations, we can calculate the maximal probability of getting 3 \emph{bar}s after next spin (see Table ~\ref{tab:diff}). The maximal probability in the generating model for different $N$ are the same since there is still one reel is that has not been spun. We can observe that conditional probabilities in learned models are quite different from the ones in generating models.



\begin{table}[htbp]
  \caption{conditional probability}
 \centering
 \begin{tabular}{|c|c|c|c|c|c|}
\hline
 & real & N=4 & N=6 & N=8 & N=10\\
\hline 
 $P(3\times \text{\emph{bar}s} \mid C_1)$& 0.30 & 0.0714 & 0.0356 & 0.0327 & 0.0450\\
\hline
 $P(3\times \text{\emph{bar}s} \mid C_2)$& 0.04 & 0.0551 & 0.0659 & 0.0934 & 0.0701\\
\hline
 $P(3\times \text{\emph{bar}s} \mid C_3)$ & 0.30 & 0.0940 & 0.0835 & 0.0874 & 0.0885\\
\hline
\end{tabular}
\label{tab:diff}
\end{table}

%
%
%

\section{Conclusion}
\label{sec:conclusion}
In this paper, we have proposed the \ioalergia\ algorithm for learning deterministic labeled Markov processes (DLMDPs). Given sequences of alternating input and output symbols, the algorithm can automatically construct a model, for the reactive system under observation, and we have similar convergence result of the \ioalergia\ algorithm as given in~\cite{MCJN+11} for deterministic Markov chain models. The algorithm is empirically analyzed using a case study based on slot machines. The learning results are evaluated by comparing in terms of  PLTL properties and maximal expected rewards of both the learned model with the known generating models as well as the accuracy of optimal actions derived from the learned models.

Compared to the learning algorithm for deterministic automata~\cite{RafSteBer05}, further research is required to make the learning algorithm that suitable for routine use. In addition to empirically demonstrating the learned model is a good approximation, measuring the distance between the learned model and the generating model will be part of our future work. For compositional systems, this learning approach could be extended to learn models for each individual component from the observed interaction among components. Moreover, the approach for learning DLMDP could be refined by \emph{active learning} techniques that take advantage of interactive data acquisition.

\bibliographystyle{eptcs}

\begin{thebibliography}{10}
\providecommand{\bibitemdeclare}[2]{}
\providecommand{\surnamestart}{}
\providecommand{\surnameend}{}
\providecommand{\urlprefix}{Available at }
\providecommand{\url}[1]{\texttt{#1}}
\providecommand{\href}[2]{\texttt{#2}}
\providecommand{\urlalt}[2]{\href{#1}{#2}}
\providecommand{\doi}[1]{doi:\urlalt{http://dx.doi.org/#1}{#1}}
\providecommand{\bibinfo}[2]{#2}

\bibitemdeclare{inproceedings}{AartsV10}
\bibitem{AartsV10}
\bibinfo{author}{Fides \surnamestart Aarts\surnameend} \&
  \bibinfo{author}{Frits~W. \surnamestart Vaandrager\surnameend}
  (\bibinfo{year}{2010}): \emph{\bibinfo{title}{Learning I/O Automata}}.
\newblock In: {\sl \bibinfo{booktitle}{CONCUR}}, pp. \bibinfo{pages}{71--85}.
\newblock \urlprefix\url{http://dx.doi.org/10.1007/978-3-642-15375-4_6}.

\bibitemdeclare{article}{Angluin87}
\bibitem{Angluin87}
\bibinfo{author}{D.~\surnamestart Angluin\surnameend} (\bibinfo{year}{1987}):
  \emph{\bibinfo{title}{Learning regular sets from queries and
  counterexamples}}.
\newblock {\sl \bibinfo{journal}{Information and Computation}}
  \bibinfo{volume}{75}, pp. \bibinfo{pages}{87--106}.

\bibitemdeclare{book}{BK2008}
\bibitem{BK2008}
\bibinfo{author}{Christel \surnamestart Baier\surnameend} \&
  \bibinfo{author}{Joost-Pieter \surnamestart Katoen\surnameend}
  (\bibinfo{year}{2008}): \emph{\bibinfo{title}{Principles of model checking}}.
\newblock \bibinfo{publisher}{MIT Press}.

\bibitemdeclare{inproceedings}{BFHH11}
\bibitem{BFHH11}
\bibinfo{author}{J.~\surnamestart Bogdoll\surnameend},
  \bibinfo{author}{L.~M.~F. \surnamestart Fioriti\surnameend},
  \bibinfo{author}{A.~\surnamestart Hartmanns\surnameend} \&
  \bibinfo{author}{H.~\surnamestart Hermanns\surnameend}
  (\bibinfo{year}{2011}): \emph{\bibinfo{title}{Partial Order Methods for
  Statistical Model Checking and Simulation}}.
\newblock In: {\sl \bibinfo{booktitle}{FMOODS/FORTE}}, pp.
  \bibinfo{pages}{59--74}.
\newblock \urlprefix\url{http://dx.doi.org/10.1007/978-3-642-21461-5_4}.

\bibitemdeclare{inproceedings}{CarOnc94}
\bibitem{CarOnc94}
\bibinfo{author}{R.~C. \surnamestart Carrasco\surnameend} \&
  \bibinfo{author}{J.~\surnamestart Oncina\surnameend} (\bibinfo{year}{1994}):
  \emph{\bibinfo{title}{Learning Stochastic Regular Grammars by Means of a
  State Merging Method}}.
\newblock In: {\sl \bibinfo{booktitle}{ICGI}}, pp. \bibinfo{pages}{139--152}.
\newblock \urlprefix\url{http://dx.doi.org/10.1007/3-540-58473-0_144}.

\bibitemdeclare{inproceedings}{CMJN+12}
\bibitem{CMJN+12}
\bibinfo{author}{Y.~\surnamestart Chen\surnameend},
  \bibinfo{author}{H.~\surnamestart Mao\surnameend},
  \bibinfo{author}{M.~\surnamestart Jaeger\surnameend}, \bibinfo{author}{T.~D.
  \surnamestart Nielsen\surnameend}, \bibinfo{author}{K.~G. \surnamestart
  Larsen\surnameend} \& \bibinfo{author}{B.~\surnamestart Nielsen\surnameend}
  (\bibinfo{year}{2012}): \emph{\bibinfo{title}{Learning Markov Models for
  Stationary System Behaviors}}.
\newblock In: {\sl \bibinfo{booktitle}{NFM}}, pp. \bibinfo{pages}{216--230}.
\newblock \urlprefix\url{http://dx.doi.org/10.1007/978-3-642-28891-3_22}.

\bibitemdeclare{inproceedings}{HigTho00}
\bibitem{HigTho00}
\bibinfo{author}{Colin \surnamestart de~la Higuera\surnameend} \&
  \bibinfo{author}{Franck \surnamestart Thollard\surnameend}
  (\bibinfo{year}{2000}): \emph{\bibinfo{title}{Identification in the Limit
  with Probability One of Stochastic Deterministic Finite Automata}}.
\newblock In: {\sl \bibinfo{booktitle}{ICGI}}, pp. \bibinfo{pages}{141--156}.
\newblock \urlprefix\url{http://dx.doi.org/10.1007/978-3-540-45257-7_12}.

\bibitemdeclare{book}{Higuera10}
\bibitem{Higuera10}
\bibinfo{author}{Colin~{de la} \surnamestart Higuera\surnameend}
  (\bibinfo{year}{2010}): \emph{\bibinfo{title}{Grammatical Inference ---
  Learning Automata and Grammars}}.
\newblock \bibinfo{publisher}{Cambridge University Press}.

\bibitemdeclare{inproceedings}{Jansen02b}
\bibitem{Jansen02b}
\bibinfo{author}{D.~N. \surnamestart Jansen\surnameend} (\bibinfo{year}{2002}):
  \emph{\bibinfo{title}{Probabilistic {UML} Statecharts for Specification and
  Verification a Case Study}}.
\newblock In: {\sl \bibinfo{booktitle}{Critical Systems Development with UML --
  Proc. of the UML'02 workshop}}, pp. \bibinfo{pages}{121--132}.

\bibitemdeclare{inproceedings}{KNP11}
\bibitem{KNP11}
\bibinfo{author}{M.~\surnamestart Kwiatkowska\surnameend},
  \bibinfo{author}{G.~\surnamestart Norman\surnameend} \&
  \bibinfo{author}{D.~\surnamestart Parker\surnameend} (\bibinfo{year}{2011}):
  \emph{\bibinfo{title}{{PRISM} 4.0: Verification of Probabilistic Real-time
  Systems}}.
\newblock In: {\sl \bibinfo{booktitle}{CAV}}, {\sl \bibinfo{series}{LNCS}}
  \bibinfo{volume}{6806}, \bibinfo{publisher}{Springer}, pp.
  \bibinfo{pages}{585--591}.

\bibitemdeclare{inproceedings}{Legay10}
\bibitem{Legay10}
\bibinfo{author}{A.~\surnamestart Legay\surnameend},
  \bibinfo{author}{B.~\surnamestart Delahaye\surnameend} \&
  \bibinfo{author}{S.~\surnamestart Bensalem\surnameend}
  (\bibinfo{year}{2010}): \emph{\bibinfo{title}{Statistical Model Checking: An
  Overview}}.
\newblock In: {\sl \bibinfo{booktitle}{RV}}, pp. \bibinfo{pages}{122--135}.
\newblock \urlprefix\url{http://dx.doi.org/10.1007/978-3-642-16612-9_11}.

\bibitemdeclare{inproceedings}{Leucker06}
\bibitem{Leucker06}
\bibinfo{author}{Martin \surnamestart Leucker\surnameend}
  (\bibinfo{year}{2006}): \emph{\bibinfo{title}{Learning Meets Verification}}.
\newblock In: {\sl \bibinfo{booktitle}{FMCO}}, pp. \bibinfo{pages}{127--151}.
\newblock \urlprefix\url{http://dx.doi.org/10.1007/978-3-540-74792-5_6}.

\bibitemdeclare{inproceedings}{MCJN+11}
\bibitem{MCJN+11}
\bibinfo{author}{H.~\surnamestart Mao\surnameend},
  \bibinfo{author}{Y.~\surnamestart Chen\surnameend},
  \bibinfo{author}{M.~\surnamestart Jaeger\surnameend}, \bibinfo{author}{T.~D.
  \surnamestart Nielsen\surnameend}, \bibinfo{author}{K.~G. \surnamestart
  Larsen\surnameend} \& \bibinfo{author}{B.~\surnamestart Nielsen\surnameend}
  (\bibinfo{year}{2011}): \emph{\bibinfo{title}{Learning Probabilistic Automata
  for Model Checking}}.
\newblock In: {\sl \bibinfo{booktitle}{QEST}}, pp. \bibinfo{pages}{111--120}.
\newblock
  \urlprefix\url{http://doi.ieeecomputersociety.org/10.1109/QEST.2011.21}.

\bibitemdeclare{inproceedings}{RafSteBer05}
\bibitem{RafSteBer05}
\bibinfo{author}{H.~\surnamestart Raffelt\surnameend} \&
  \bibinfo{author}{B.~\surnamestart Steffen\surnameend} (\bibinfo{year}{2006}):
  \emph{\bibinfo{title}{LearnLib: A Library for Automata Learning and
  Experimentation}}.
\newblock In: {\sl \bibinfo{booktitle}{FASE}}, pp. \bibinfo{pages}{377--380}.
\newblock \urlprefix\url{http://dx.doi.org/10.1007/11693017_28}.

\bibitemdeclare{article}{RS1996}
\bibitem{RS1996}
\bibinfo{author}{D.~\surnamestart Ron\surnameend},
  \bibinfo{author}{Y.~\surnamestart Singer\surnameend} \&
  \bibinfo{author}{N.~\surnamestart Tishby\surnameend} (\bibinfo{year}{1996}):
  \emph{\bibinfo{title}{The Power of Amnesia: Learning Probabilistic Automata
  with Variable Memory Length}}.
\newblock {\sl \bibinfo{journal}{Machine Learning}}
  \bibinfo{volume}{25}(\bibinfo{number}{2-3}), pp. \bibinfo{pages}{117--149}.
\newblock \urlprefix\url{http://dx.doi.org/10.1023/A:1026490906255}.

\bibitemdeclare{inproceedings}{SVA04a}
\bibitem{SVA04a}
\bibinfo{author}{K.~\surnamestart Sen\surnameend},
  \bibinfo{author}{M.~\surnamestart Viswanathan\surnameend} \&
  \bibinfo{author}{G.~\surnamestart Agha\surnameend} (\bibinfo{year}{2004}):
  \emph{\bibinfo{title}{Learning Continuous Time Markov Chains from Sample
  Executions}}.
\newblock In: {\sl \bibinfo{booktitle}{QEST}}, pp. \bibinfo{pages}{146--155}.
\newblock
  \urlprefix\url{http://doi.ieeecomputersociety.org/10.1109/QEST.2004.10014}.

\bibitemdeclare{article}{Stoelinga02}
\bibitem{Stoelinga02}
\bibinfo{author}{Mari{\"e}lle \surnamestart Stoelinga\surnameend}
  (\bibinfo{year}{2002}): \emph{\bibinfo{title}{An Introduction to
  Probabilistic Automata}}.
\newblock {\sl \bibinfo{journal}{Bulletin of the EATCS}} \bibinfo{volume}{78},
  pp. \bibinfo{pages}{176--198}.

\bibitemdeclare{book}{TanSteKum06}
\bibitem{TanSteKum06}
\bibinfo{author}{P.-N. \surnamestart Tan\surnameend},
  \bibinfo{author}{M.~\surnamestart Steinbach\surnameend} \&
  \bibinfo{author}{V.~\surnamestart Kumar\surnameend} (\bibinfo{year}{2006}):
  \emph{\bibinfo{title}{Introduction to Data Mining}}.
\newblock \bibinfo{publisher}{Addison Wesley}.

\bibitemdeclare{article}{Hoeffding63}
\bibitem{Hoeffding63}
\bibinfo{author}{H.~\surnamestart Wassily\surnameend} (\bibinfo{year}{1963}):
  \emph{\bibinfo{title}{Probability Inequalities for Sums of Bounded Random
  Variables}}.
\newblock {\sl \bibinfo{journal}{Journal of the American Statistical
  Association}} \bibinfo{volume}{58}(\bibinfo{number}{58}), pp.
  \bibinfo{pages}{13--30}.
\newblock \urlprefix\url{http://dx.doi.org/10.2307/2282952}.

\bibitemdeclare{inproceedings}{YS02}
\bibitem{YS02}
\bibinfo{author}{H.~L.~S. \surnamestart Younes\surnameend} \&
  \bibinfo{author}{R.~G. \surnamestart Simmons\surnameend}
  (\bibinfo{year}{2002}): \emph{\bibinfo{title}{Probabilistic Verification of
  Discrete Event Systems Using Acceptance Sampling}}.
\newblock In: {\sl \bibinfo{booktitle}{CAV}}, pp. \bibinfo{pages}{223--235}.
\newblock \urlprefix\url{http://dx.doi.org/10.1007/3-540-45657-0_17}.

\end{thebibliography}

\end{document}